\documentclass[journal]{IEEEtran}
\usepackage{tabularx,booktabs}
\newcolumntype{C}{>{\centering\arraybackslash}X} 
\IEEEoverridecommandlockouts
\IEEEpubid{\makebox[\columnwidth]{} \hspace{\columnsep}\makebox[\columnwidth]{ }}
\input epsf
\usepackage[pdftex]{graphicx}
\usepackage{epstopdf}
\usepackage{multirow}
\usepackage{amsmath}
\usepackage[draft]{hyperref}
\usepackage{url}
\usepackage{verbatim} 
\usepackage{subfigure}  
\usepackage{cite}
\usepackage{adjustbox}
\usepackage{longtable}
\usepackage{lscape}

\ifCLASSINFOpdf
\else
\fi
\begin{document}

\title{A CODECO Case Study and Initial Validation for Edge Orchestration of Autonomous Mobile Robots}

\author{\IEEEauthorblockN{Hongyu Zhu, Tina Samizadeh, Rute C. Sofia}\\
\IEEEauthorblockA{fortiss - research Institute of the Free State of Bavaria associated with 
\\
the Technical University of Munich (TUM)
} 
 }

\maketitle 

\begin{abstract}
Autonomous Mobile Robots (AMRs) increasingly adopt containerized micro-services across the Edge-Cloud continuum. While Kubernetes is the de-facto orchestrator for such systems, its assumptions—stable networks, homogeneous resources, and ample compute capacity do not fully hold in mobile, resource-constrained robotic environments.

The paper describes a case-study on smart-manufacturing AMR and performs an initial comparison between CODECO orchestration and standard Kubernetes using a controlled Kubernetes-in-Docker (KinD) environment. Metrics include pod deployment and deletion times, CPU and memory usage, and inter-pod data rates. The observed results indicate that CODECO offers reduced CPU consumption and more stable communication patterns, at the cost of modest memory overhead (~10–15\%) and slightly increased pod lifecycle latency due to secure overlay initialization.
 \end{abstract}
\maketitle
\section{Introduction}
Industrial IoT (IIoT) applications increasingly adopt micro-service architectures and deploy workloads across the Edge-Cloud continuum using container technologies such as Docker\footnote{\href{https://www.docker.com/}{https://www.docker.com/}}. Deployment and lifecycle management are typically handled by container orchestrators, with Kubernetes being the de facto standard.

Kubernetes provides declarative configuration, automated scaling, and robust availability mechanisms that make it highly effective in cloud data-centers. However, its design assumptions, namely, the existence of relatively stable networks, abundant compute resources, and largely static infrastructure, do not fully hold in Edge-Edge and Edge-Cloud environments. In such settings, resources can be constrained and heterogeneous. The nodes are often mobile and interconnected through wireless and cellular technologies. Hence, conditions may vary unpredictably, as often encountered in \textit{Autonomous Mobile Robot (AMR)} deployments.

To address these challenges, extensions to Kubernetes can integrate context-awareness, Quality of Service (QoS) requirements, and even Quality of Experience (QoE) requirements.

In this context, the Cognitive Decentralized Edge-Cloud Orchestration (CODECO) framework extends Kubernetes with mechanisms for telemetry-driven orchestration in resource-constrained dynamic settings~\cite{CODECO_framework}. CODECO augments Kubernetes by considering application-level QoS, user-defined performance profiles (e.g., resilience, latency, energy efficiency), and real-time telemetry on compute, network, and data resources.

This article presents a focused case study that applies CODECO to an AMR-oriented smart-manufacturing scenario. We investigate the following research questions (RQ):

\begin{itemize}
  \item \textbf{RQ1} Can CODECO improve resource usage both at a compute and networking level when compared to baseline Kubernetes for AMR micro-service workloads?
  \item \textbf{RQ2} Which operational costs (e.g., memory overhead, pod-lifecycle latency) accompany these potential benefits?
  \item \textbf{RQ3} Under what deployment conditions do these trade-offs favor the use of CODECO in robotics?
\end{itemize}

The contributions of this article are as follows:

\begin{itemize}
  \item A self-contained overview of CODECO tailored to AMR practitioners and robotics researchers.
  \item An initial comparison between Kubernetes and Kubernetes+CODECO in a KinD-based emulator, evaluating CPU and memory usage, inter-pod data rates, and pod deployment/deletion latency.
  \item A balanced discussion of orchestration trade-offs in distributed AMR systems, supported by empirical results and a roadmap toward real-device, multi-robot, wireless-aware validation.
\end{itemize}

\subsection{Study Type and Limitations}
This work represents an \emph{initial validation and case study}. Experiments are conducted in an emulator (KinD) with a minimal AMR-like micro-service workload. Large-scale fleet scenarios, heterogeneous embedded platforms, wireless impairments, and energy-sensitive scheduling are reserved for future real-world experimentation.

\begin{figure*}[htp!]
    \centering
    \includegraphics[width=\textwidth]{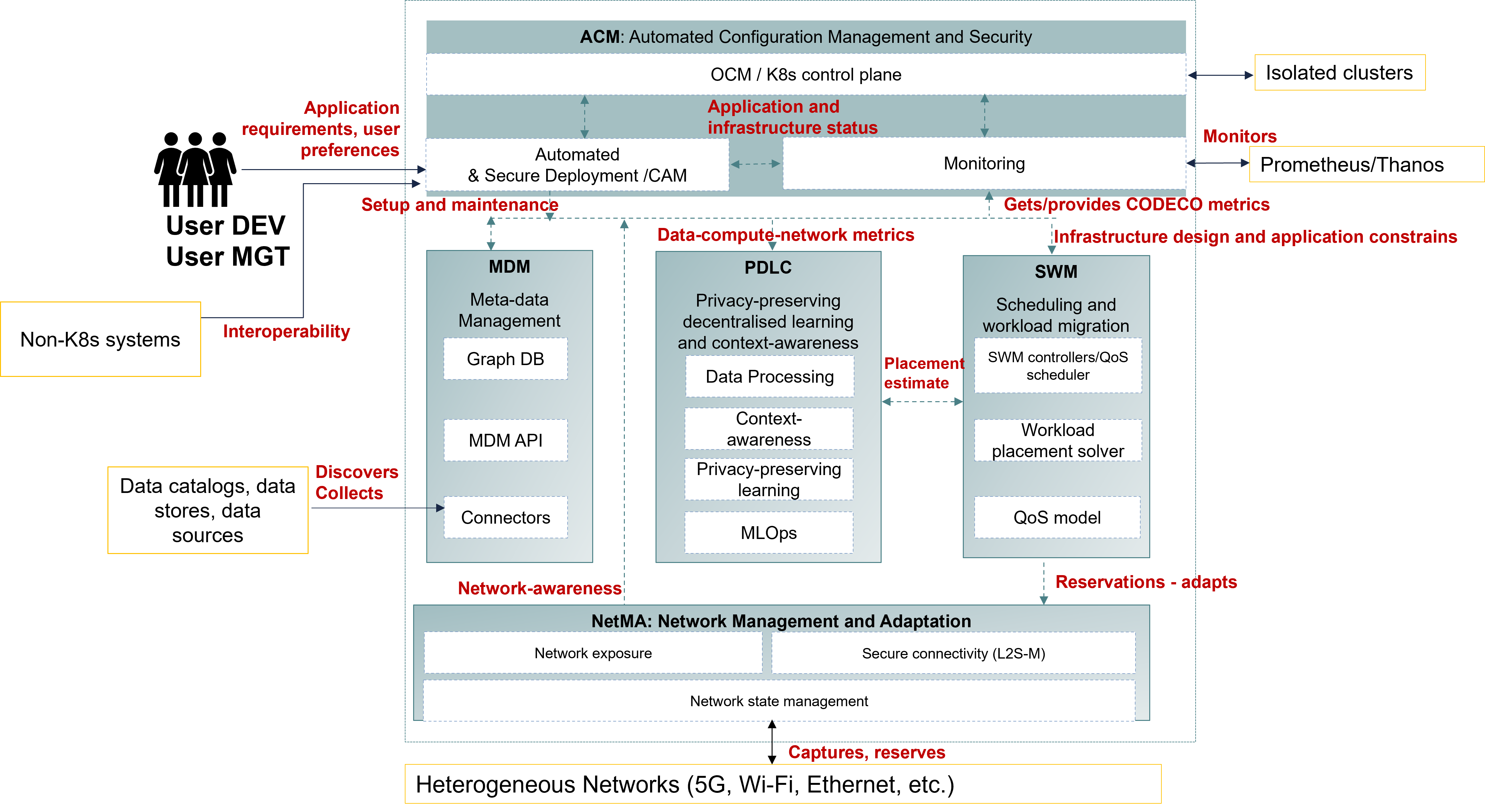} 
    \caption{The CODECO framework and its components. dashed lines represent internal interfaces based on Kubernetes APIs. Continuous arrows represent interfaces to non-Kubernetes systems, e.g., user, data catalogues. }
    \label{fig:CODECO}
\end{figure*}

\section{Edge-Cloud Container  Basics}
\label{background}

Edge, Cloud Definitions, including the notions of \textbf{far Edge} and \textbf{near Edge} follow the line of thought being driven in European initiatives, such as  EUCloudEdgeIoT.eu\footnote{\href{https://eucloudedgeiot.eu/}{https://eucloudedgeiot.eu/}}~\cite{sofia-a}. 

A \textbf{container}, per Kubernetes, is a unit that runs an application independently within a \textbf{pod}. A \textbf{pod} groups one or more tightly linked containers sharing resources. A Kubernetes \textbf{cluster} is the orchestrated environment in which pods are managed by the \textbf{user}~\textbf{user}.

Kubernetes, the leading orchestrator, offers declarative configuration, automation, and resilience. Applications run in containers grouped into pods, managed within a cluster. Kubernetes ensures scalability, load balancing, health checks, self-healing, and rolling updates. While well-suited to Cloud environments, Kubernetes faces limitations in Edge-Cloud settings, which are more dynamic, resource-constrained, and often mobile~\cite{8457916}.

\section{CODECO in a Nutshell}
\label{CODECO}

The CODECO framework represented in Figure \ref{fig:CODECO} extends Kubernetes to support dynamic, resource-constrained, and mobile computing environments such as robotics and Industrial IoT. While Kubernetes assumes reliable connectivity and abundant resources, CODECO introduces mechanisms to operate in heterogeneous, bandwidth-limited, and latency-sensitive scenarios.

CODECO provides orchestration intelligence across compute, data, and network resources, enabling applications to be deployed, migrated, and optimized across the Edge-Cloud continuum. It incorporates telemetry-driven decision-making, user-defined performance goals (e.g., latency, resilience, energy efficiency), and secure multi-node coordination. This integrated view is referred to as \textbf{data–compute–network orchestration}.

The notion of infrastructure in CODECO considers different infrastructure perspectives: the networking perspective, where the network is perceived both as the underlay and overlay infrastructure required to support the deployment of containerized applications across Edge-Cloud; the data observability perspective, bringing input on the data dependencies that micro-services may have, and that is important to consider during the application deployment and re-deployment; the computational perspective, bringing input on the suitable nodes to serve an application, taking into consideration.

 \subsection{Components and Operational Roles}
\label{subsec: components}
CODECO introduces several components that run alongside Kubernetes control-plane services. Users interact with CODECO either as application developers (DEV) or cluster operators (MGT).

The only interface of CODECO to the user, as shown in Figure \ref{fig:CODECO}, is the \textbf{Automated Configuration Management and Security (ACM)} component. ACM Installs and configures CODECO services and provides status on the application and on the infrastructure to the user via a dashboard. It processes application requirements and performance objectives defined through the \textbf{CODECO Application Model (CAM)}. The CAM is a declarative model where users specify micro-service requirements (latency, energy, resilience, network constraints) and desired performance profiles (e.g., energy-efficient or fault-tolerant operation)\footnote{Examples of CAM and respective schema are available in the CODECO Eclipse Git repositories: https://gitlab.eclipse.org/eclipse-research-labs/codeco-project/acm/-/tree/main/config/samples?ref\_type=heads}. Such specification is relevant to allow CODECO to deploy the application across Edge-Cloud, without having to check into the application. Once CODECO is installed, all CODECO components can be used. The CODECO components have been devised independently and can be used with other future open orchestration frameworks, requiring little adaptation.

The CODECO \textbf{Metadata Manager (MDM)} component provides data and system observability to the other CODECO components, treating data as an integral part of the application workload. MDM collects data usage metrics such as freshness, age of information, and brings such data to Prometheus. hence, MDM treats data flows as first-class resources, unlike plain Kubernetes.

The \textbf{Network Management and Adaptation (NetMA)} component provides network probing and awareness (latency, bandwidth, link quality, energy consumed per link). It also establishes secure, Layer-2 connectivity between pods via the L2S-M mechanism, simplifying pod-to-pod communication across distributed and mobile nodes.

The \textbf{Privacy-preserving Decentralised Learning and Context-awareness (PDLC) }component aggregates the monitored infrastructure metrics to compute a node weight for each performance profile. It also provides learning-based stability estimation and can run across control-plane and worker nodes to support decentralised intelligence. Firstly, it provides an aggregated cost view of a specific target performance profile for the available infra-structure which can be used by other components and is currently being considered in SWM to further define the optimal workload placement. Secondly, it provides an estimate on the overall system stability based on privacy-preserving decentralised learning approaches. PDLC is currently envisaged to operate on both Kubernetes master and worker nodes.

The \textbf{Seamless Workload Migration (SWM)} component performs placement and migration decisions based on CAM and runtime metrics collected by ACM (application and user), MDM (data), and NetMA (network). Unlike Kubernetes’ filter-and-score placement approach, SWM uses a graph-based optimization approach that considers the application QoS/QoE requirements defined in the CAM.

Therefore, to be able to achieve a flexible orchestration, CODECO counts with monitoring across the three different Edge-Cloud infrastructure perspectives. NetMA monitors the networking infrastructure; MDM monitors the data infrastructure; ACM monitors the system (computational nodes) infrastructure based on the Kubernetes metrics server Prometheus. This approach allows to explore the integration of user defined metrics as well as of providing the collected metrics of CODECO to future orchestration frameworks.

The CODECO framework is composed of modular micro-services: each CODECO component has been devised following a modular approach, integrating different sub-components that are expected to be built as one or more independent (dockerized) micro-services.

CODECO’s architecture makes it suitable for robotics applications orchestration. AMRs typically run perception and control tasks locally while relying on the Cloud for heavy computation. Kubernetes alone does not adapt to connectivity variability, energy constraints, or real-time mobility. CODECO fills this gap through telemetry-driven decisions, stateful migration, and secure networking across heterogeneous nodes. 

\section{Orchestrating AMR Applications with Kubernetes and CODECO}
\label{roboticsorchestration}
\begin{table*}[htp!]
\centering
\caption{Comparison of plain Kubernetes and CODECO for AMR application orchestration.}
\label{tab:Kubernetes_codeco_comparison}
\begin{tabular}{p{4cm}p{5cm}p{5cm}}
\toprule
\textbf{Feature} & \textbf{Kubernetes} & \textbf{CODECO} \\
\midrule
\textit{Real-Time Performance} & Not optimized for real-time constraints; may introduce latency & Uses AI-based recommendations for the scheduling of robotic workloads \\
\textit{Network Adaptability} & Does not handle intermittent connectivity well & Adapts dynamically to changing network conditions \\
\textit{Resource Optimization for Edge Devices} & High resource overhead; requires lightweight alternatives (K3s, KubeEdge) & Optimized workload allocation based on CPU, memory, battery, energy expenditure, network state \\
\textit{Security} \& Privacy & Provides RBAC \& Network Policies & Privacy-Preserving Learning, secure connectivity, data compliance \\
\textit{Data Management \& Synchronization} & Manages persistent volumes but does not optimize robotic data flow & Ensures efficient data synchronization \\
\textit{Ease of Deployment \& Configuration} & Requires manual tuning; DevOps-heavy & Automates Kubernetes configurations, reducing manual setup \\
\textit{Scalability \& Flexibility} & Scalable across Cloud \& Edge but lacks AI-driven optimization & AI-driven orchestration expected to increase automated behaviour of AMR fleets \\
\textit{Self-Healing \& Fault Tolerance} & Can restart pods but lacks AI-based failure prediction & Predictive self-healing and AI-based anomaly detection prevent failures \\
\bottomrule
\end{tabular}
\end{table*}

At the Edge, AMRs rely on systems such as ROS~2 or custom middleware to manage onboard computation\footnote{\url{https://docs.ros.org/en/humble/index.html}}. Real-time tasks like motion control, Simultaneous Localization and Mapping (SLAM), and perception often run on embedded devices. Micro-services may be containerized for modularity, while time-critical decisions such as obstacle avoidance remain local. ROS~2’s flexibility has made it a common foundation for robotics, supporting scalable deployment with Docker and communication via the Data Distribution Service (DDS)\footnote{\href{https://docs.ros.org/en/humble/Installation/DDS-Implementations.html}{DDS Implementations}}~\cite{7743223}. DDS supports Publish-Subscribe (PubSub) communication between ROS 2 packages (containerized micro-services). Hence, Kubernetes simplifies the required management of ROS~2 micro-services by automating deployment and coordination, but it is not optimized for real-time robotics. Limitations include DDS latency, scheduling delays, networking overhead, and constrained Edge resources. Challenges are amplified by wireless connectivity, high-bandwidth sensors, and lack of awareness of robotics-specific metrics such as battery level.

CODECO addresses these issues via its QoS-aware scheduling, dynamic workload migration, and adaptive redistribution under failures or energy constraints. It automates Kubernetes configurations, easing the burden on engineers. Using the CODECO Application Model, robotics developers can specify mission goals and performance targets, while CODECO manages deployment and reconfiguration. In dynamic environments such as factories, CODECO supports runtime adjustments and efficient workload offloading when nodes are energy-depleted or poorly connected.

A comparison of the plain Kubernetes vs. Kubernetes with CODECO orchestration for AMR applications is provided in Table \ref{tab:Kubernetes_codeco_comparison}.

\section{Case Study}
\label{P5 Use-case}
This section describes the case study used to perform an initial validation of CODECO in the context of AMR application orchestration in an indoor smart Manufacturing environment. The objective of the case study design is to understand the benefits of CODECO in terms of \textbf{stateful micro-service migration}, resource efficiency, and communication stability under realistic constraints, reflecting energy variability, wireless fluctuations, and dynamic topology changes.

The case-study aims at answering the following research questions: 
\begin{itemize}
    \item  How efficiently can CODECO perform stateful micro-service migration on AMRs, while maintaining service continuity?
    \item  Does CODECO improve resource utilization and energy-aware scheduling compared to default Kubernetes orchestration?
     \item   Can CODECO contribute to communication stability under variable AMR conditions, such as battery depletion?
\end{itemize}

\begin{figure*}[htp!]
    \centering
    \includegraphics[width=\textwidth]{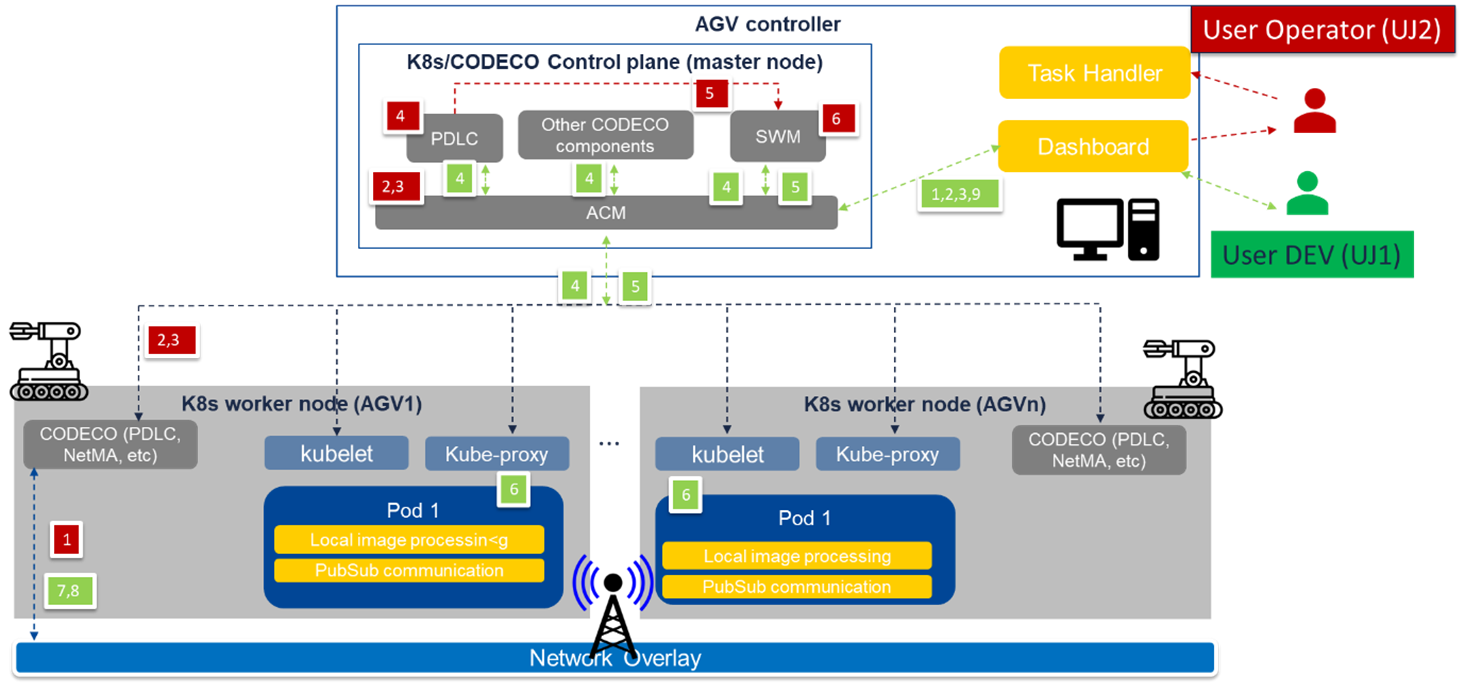} 
    \caption{CODECO for AMR fleet orchestration: deployment and runtime management overview. Red arrows and boxes relate with a user story for an AMR manager (application runtime management). Green arrows and boxes correspond to a user story for the application initial setup, usually handled by an AMR application developer.\label{fig:CODECO_P5}}
\end{figure*}

Figure~\ref{fig:CODECO_P5} illustrates the system concept. Two AMRs (Kubernetes worker nodes, AGV1 and AGV2) execute SLAM, navigation, and control tasks, requiring low latency, predictable communication, and workload offloading to edge controllers when resources fluctuate. A controller node (AGV controller) containers the Kubernetes control plane and CODECO components, including the new scheduler. On the AMRs (worker nodes), CODECO runs monitoring components. A full description of the conceptual use-case that is the basis for this case study is available in~\cite{raquel_sousa_ed_2023_8143800}.

The current case study is being implemented on a testbed based on ROS~2 micro-services orchestrated by Kubernetes and CODECO, deployed in a hybrid setup combining physical robots and emulator-based clusters. The codebase and Docker images are open-sourced\footnote{\href{https://gitlab.eclipse.org/eclipse-research-labs/codeco-project/use-cases/p5-amr-manufacturing}{https://gitlab.eclipse.org/eclipse-research-labs/codeco-project/use-cases/p5-amr-manufacturing}} for reproducibility\footnote{The current P5 use-case is being deployed in Munich, in the fortiss IIoT Lab. An open testbed is available for visit}.

Two specific user-stories are considered, detailed next.

\subsection{AMR Application Deployment Setup}
A developer (DEV, green, Figure \ref{fig:CODECO_P5}) deploys an application across an AMR fleet by accessing the CODECO ACM GUI (1) and defining requirements such as latency, fleet size, communication type (5G/Wi-Fi), and resilience levels (2,3). These inputs form a CAM, guiding deployment. CODECO stores the CAM (4) and exposes it via Kubernetes interfaces. The Scheduler (SWM) initiates placement (5), deploying to a default cluster that expands as new nodes appear (6). CODECO manages secure connectivity (7) and configures routes as needed (8). The developer can monitor and adjust deployments through the dashboard (9).

\subsection{AMR Application Runtime Management}
This user story (MGT) focuses on runtime management to support autonomous navigation in constrained indoor spaces. During the application runtime, CODECO monitors the infrastructure resources both at a computational and networking level. The fleet controller maintains a global view of data, compute, and network (1), updated by CODECO components (2,3). The AI-driven performance adaptation module (PDLC) provides recommendations towards placement decisions, i.e., it provides suitable nodes weights for each node in a cluster based on existing resources such as CPU, memory, latency, and also target performance profiles selected by DEV in the CAM. SWM re-scheduling occurs when a change to the infrastructure occurs, following the Kubernetes principles. An AI engine (PDLC) recommends re-scheduling based on selected performance profiles (4). SWM applies these recommendations (5) and adapts the infrastructure (6). For example, if an AMR is about to exhaust its battery, its micro-services are migrated to another suitable AMR automatically chosen by CODECO.  For instance, if an AMR battery drops below a threshold, or wireless quality degrades, some micro-services being supported by the AMR will be migrated to another robot or Edge node while preserving application state. This capability supports uninterrupted operation and resilience in dynamic, resource-limited environments.

\subsection{AMR Application Workload Design}
\label{subsec:workload}
\begin{figure}[htp!]
    \centering
    \includegraphics[width=0.7\columnwidth]{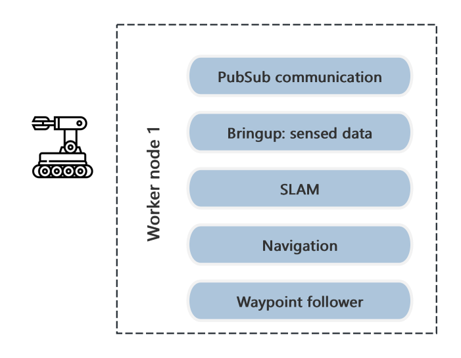} 
    \caption{The AMR application and its micro-services.}
    \label{fig:MAR-app}
\end{figure}
For this case-study, we have crated an AMR application composed of several micro-service. Each micro-service corresponds to a Docker container:

\begin{itemize}
    \item \textbf{Bringup} initializes motors, sensors, base ROS~2 stack, and DDS communication.
    \item \textbf{SLAM} runs Google Cartographer\footnote{\href{https://google-cartographer-ros.readthedocs.io/en/latest/}{Cartographer ROS2}} for mapping and localization.
    \item \textbf{Navigation (Nav2)} tracks waypoints and obstacle avoidance.
    \item \textbf{Continuous SLAM}  supports runtime migration using Kubernetes persistent volumes (PV/PVC).
\end{itemize}

Each AMR runs \textbf{bringup} locally, while compute-intensive \textbf{SLAM} and \textbf{Navigation} execute on an Edge controller or is moved (migrate) between robots. It is highlighted that CODECO supports stateful workload migration. 

ROS~2 topics transmit laser scans, odometry, and control messages using DDS PubSub.
The workload implementation follows a reproducible containerized structure provided in Table \ref{tab:app_files}\footnote{https://gitlab.eclipse.org/eclipse-research-labs/codeco-project/use-cases/p5-amr-manufacturing}. Public container images are hosted at DockerHub\footnote{https://hub.docker.com/u/hecodeco}.

\begin{table}[h!]
\caption{Application structure overview}
\label{tab:app_files}
\renewcommand{\arraystretch}{1.2}

\begin{tabularx}{\linewidth}{lX}
\toprule
\textbf{Folder} & \textbf{Description} \\
\midrule
\texttt{docker/} & Dockerfiles for ROS~2 micro-services (ARM/x86) \\
\texttt{maps/} & Indoor laboratory maps for repeatable experiments \\
\texttt{src/} & Autonomous mobile robot (AMR) application and supporting micro-services \\
\texttt{codeco\_kind/} & KiND deployment configuration and cluster automation scripts \\
\texttt{api/} & Dashboard visualisation and user interface components \\
\bottomrule
\end{tabularx}
\end{table}

\subsection{Continuous SLAM and Stateful Migration}
\label{subsec:continuous_slam}
To support uninterrupted mission execution, Kubernetes Persistent Volumes (PV) storage ensures map and state consistency when the AMR application micro-services are required to be re-scheduled. CODECO triggers migrations based on telemetry, reducing developer effort and enhancing fleet resilience.

\section{Case Study Experimental Methods}
\label{Experimentation}

This section describes the case study experimental setup. The objective is to assess orchestration overhead, compute efficiency, communication stability, and stateful workload migration based on the proposed case-study, comparing \textbf{Kubernetes vs.\ Kubernetes+CODECO}.

The evaluation is based on: a controlled environment based on Kubernetes-in-Docker (KinD) for reproducible benchmarking and a local testbed as \textbf{demonstration of capability}.

\subsection{KinD Experimental Environment}
Table~\ref{tab:environment} summarizes the machine configuration used for the KinD emulator experiments. Prometheus and Grafana were deployed for telemetry collection and real-time visualization, capturing both compute and network-level metrics.

\begin{table}[h!]
\caption{System specifications for KinD experiments.}
\centering
\scriptsize
\begin{tabular}{ll}
\toprule
\textbf{Component} & \textbf{Specification} \\
\midrule
Laptop & Lenovo ThinkPad T470p \\
RAM & 16\,GiB \\
CPU & Intel\textsuperscript{\textregistered} i7-7700HQ (8 cores, 2.80\,GHz) \\
Arch. & 64-bit \\
OS & Ubuntu 22.04.5 LTS \\
K8s Emulator & KinD v0.20.0 (Docker backend) \\
\bottomrule
\end{tabular}
\label{tab:environment}
\end{table}

The AMR application is based on ROS~2 Humble with DDS Pub/Sub, Kubernetes v1.28, Docker v24.0, and the CODECO OSS for single cluster operation\footnote{\url{https://gitlab.eclipse.org/eclipse-research-labs/codeco-project/use-cases/p5-amr-manufacturing/-/tree/main/src/turtlebot3?ref\_type=heads}}. The same manifests were deployed across plain Kubernetes and Kubernetes+CODECO to ensure comparability.

\subsection{Workload Description}
The test workload reflects a realistic AMR pipeline and consists of four ROS~2 micro-services: \emph{bringup}, \emph{SLAM}, \emph{navigation}, and \emph{continuous SLAM}, as described before. In the KinD setup, SLAM and navigation run as computational workloads, while the real TurtleBot3 AMRs validate container execution and stateful handover.

To generate consistent robotic traffic patterns, a ROS~2 talker (publisher) and listener (subscriber) pair exchange sensor-like messages across different worker nodes, simulating distributed AMR sensing and control.

\subsection{Evaluation Metrics}
The metrics being evaluated are:
\begin{itemize}
    \item \textbf{CPU usage}:  a query to Prometheus has been set to compute the per-second instant rate of CPU usage over a 1-minute timewindow for each container within the specified pod and namespace. The query provides a near real-time CPU usage metric as a percentage of the overall available CPU.
    \item \textbf{Memory usage}: resident memory per container (MiB). the real-time memory usage relied on a Prometheus query that returns the current memory usage per pod and is presented in mebibytes (MiB) as defined in Kubernetes.
    \item \textbf{Pod deployment time}: the pod deployment time (seconds) is computed as the difference between the time a pod is created and the time it reaches the “\textit{Ready}” state \footnote{\texttt{\detokenize{kube_pod_status_ready_time{namespace="default"} - kube_pod_created{namespace="default"}}}}. These are provided in seconds, and based on Unix
timestamping (soft timestamping).
     \item \textbf{Pod deletion time}:  The pod deletion time (seconds) measures the time taken for a pod to be completely removed after a deletion request is issued, computed via a script provided in the available code.
    \item \textbf{Transmit and receive data rate}: the data rates (Mbps) from talker to listener pods and vice-versa, respectively.
\end{itemize}

These metrics are critical for AMRs which must operate under compute limits, exhibit predictable communication, and recover workloads rapidly during mobility or energy events.

\subsection{Experimental Procedure}

Each experimental has been repeated five times and run follows these steps:

\begin{enumerate}
    \item Deploy the AMR application in a KinD cluster representing the proposed use-case.
    \item Start Prometheus monitoring (1\,s sampling).
    \item Initiate sensor-traffic generation (talker--listener).
    \item Measure CPU, memory, bandwidth, and latency.
    \item Trigger pod creation/deletion and record lifecycles.
    \item Induce a migration event (SLAM relocation) and monitor state handover.
\end{enumerate}

\subsection{KinD Emulation Scope and Limitations}
KinD provides deterministic, instrumentable conditions for fairness and reproducibility, but its falls short in terms of emulation for the AMR dynamics, sensor noise, or mobility.

Thus, the reported values characterize \emph{orchestration behavior}, not the full AMR performance. Experimentation based on the existing testbed is part of the planned future work.

\section{Results}
\label{sec:results}
\subsection{CPU and Memory Usage}
To assess computational overhead, CPU and memory consumption were monitored for the ROS~2 talker and listener pods under both orchestration approaches. CPU usage was measured using a Prometheus instant-per-second rate over a one-minute window, while memory usage was obtained directly from Kubernetes pod metrics in MiB.

\begin{figure}[ht]
    \centering
    \includegraphics[width=0.7\columnwidth]{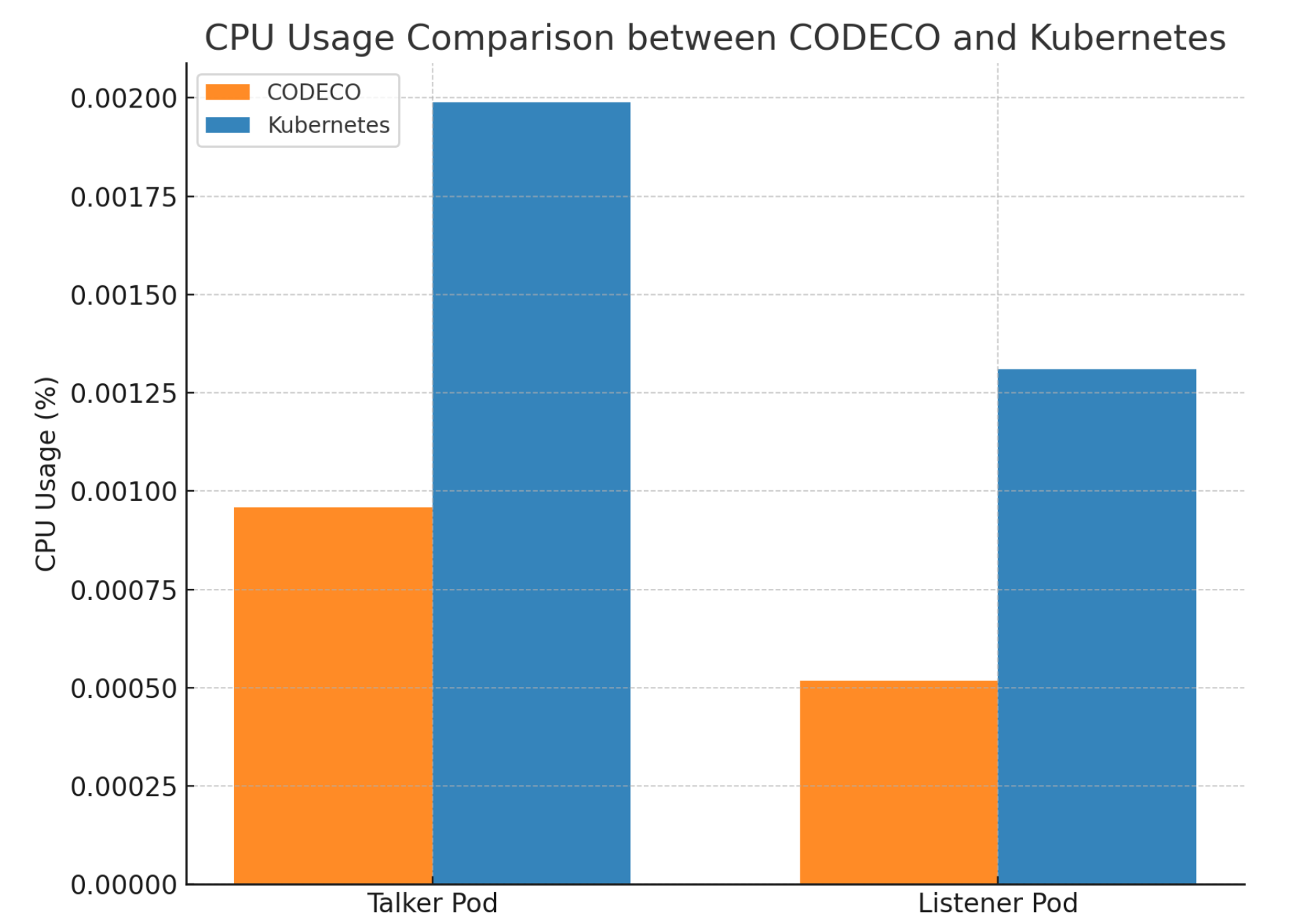}
    \caption{CPU usage under Kubernetes and Kubernetes+CODECO.}
    \label{fig:cpu}
\end{figure}

Figure~\ref{fig:cpu} shows that CODECO consistently results in \emph{lower CPU usage} in comparison to vanilla Kubernetes. Possible contributing factors to this behaviour relate with a reduced background orchestration activity and finer-grained workload placement. Further profiling is planned to better isolate the source of CPU savings.

\begin{figure}[ht]
    \centering
    \includegraphics[width=0.7\columnwidth]{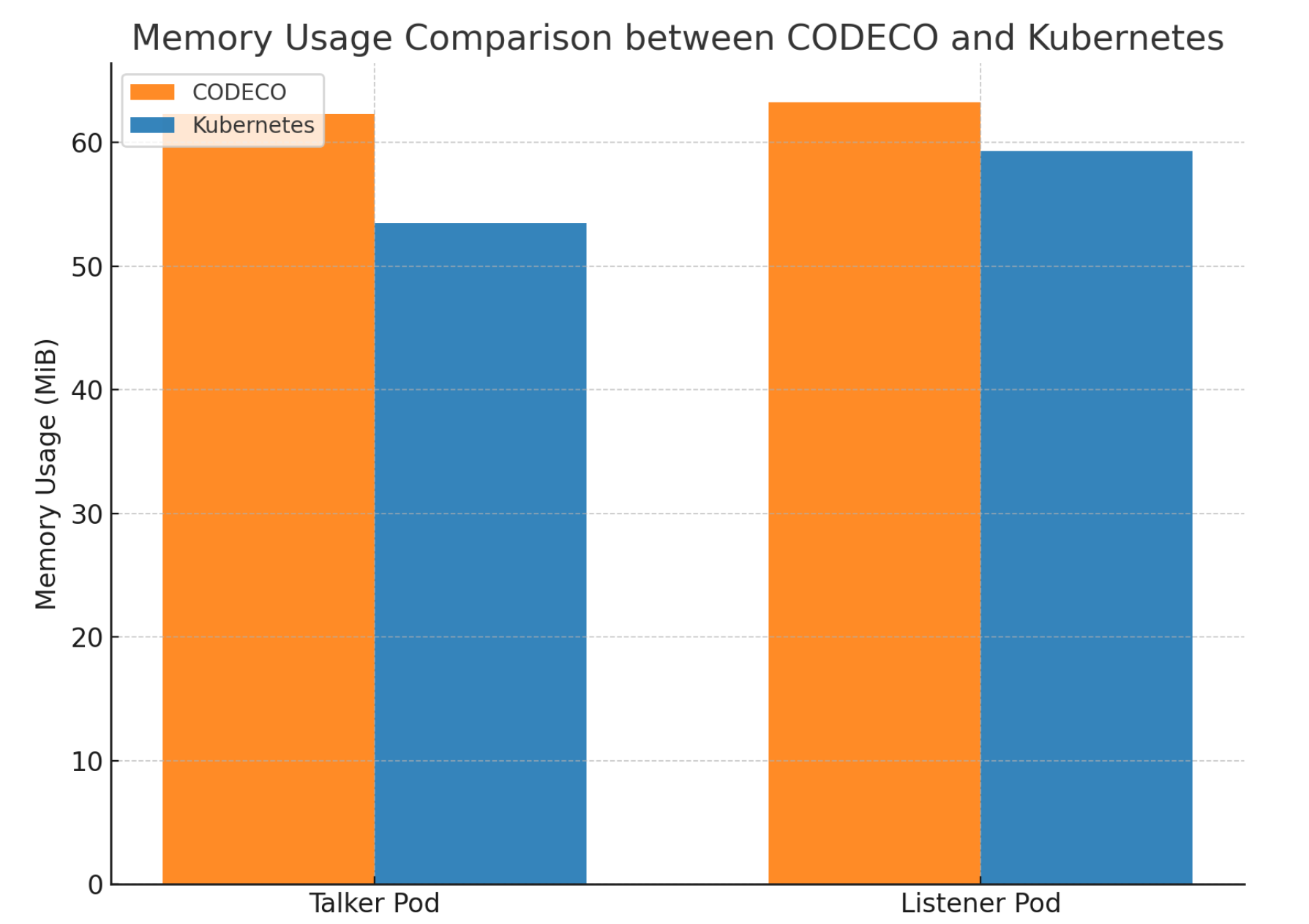}
    \caption{Memory usage under Kubernetes and Kubernetes+CODECO.}
    \label{fig:Memory}
\end{figure}

Memory usage results, provided in Figure~\ref{fig:Memory}, indicate a \textasciitilde10--15\% overhead for CODECO (e.g., 63\,MiB vs.\ 54\,MiB). This is expected due to additional services for telemetry, policy management, and state handling required for migration support. Overall, the memory impact remains modest in Kubernetes terms, relative to the capabilities added.

\subsection{Data Rates' Stability}
\begin{figure}[h!]
    \centering
    \includegraphics[width=0.9\columnwidth]{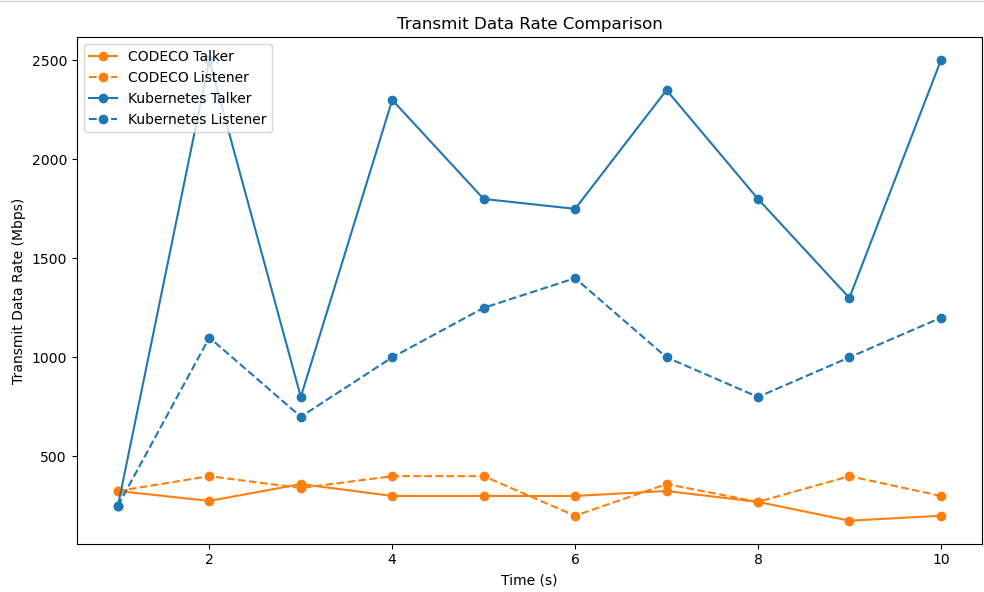}
    \caption{Transmit data rates: Kubernetes vs.\ Kubernetes+CODECO.}
    \label{fig:transmit}
\end{figure}

\begin{figure}[h!]
    \centering
    \includegraphics[width=0.9\columnwidth]{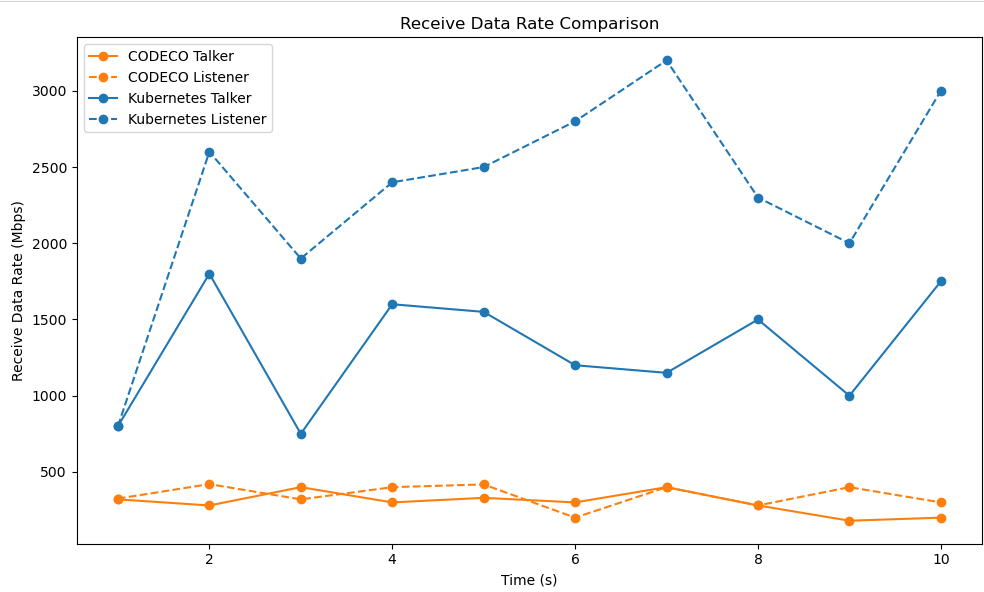}
    \caption{Receive data rates: Kubernetes vs.\ Kubernetes+CODECO.}
    \label{fig:receive}
\end{figure}

Network behavior is particularly important in  AMR environments where real-time telemetry and sensor data exchange occur over constrained wireless links. Figures~\ref{fig:transmit} and~\ref{fig:receive} respectively present the transmit and receive rates in Mbits per second (Mbps) measured between pods running on separate nodes.

Although absolute values are small due to the test workload, the trends are informative. CODECO shows noticeably \emph{more stable and lower transmit rates}, while Kubernetes displays bursts and higher peaks. Similar patterns occur for receive rates, where Kubernetes exceeds 2-3\,Gbps, whereas CODECO maintains lower, predictable ranges.

This behavior confirms that CODECO applies network-aware control aligned with QoS profiles, constraining unnecessary bursts and smoothing traffic. This characteristic is advantageous in wireless and latency-sensitive deployments, though larger-scale experiments with real mobility will be necessary to generalize the effect.

\subsection{Pod Lifecycle Times}
\begin{figure}[h!]
    \centering
    \includegraphics[width=0.7\columnwidth]{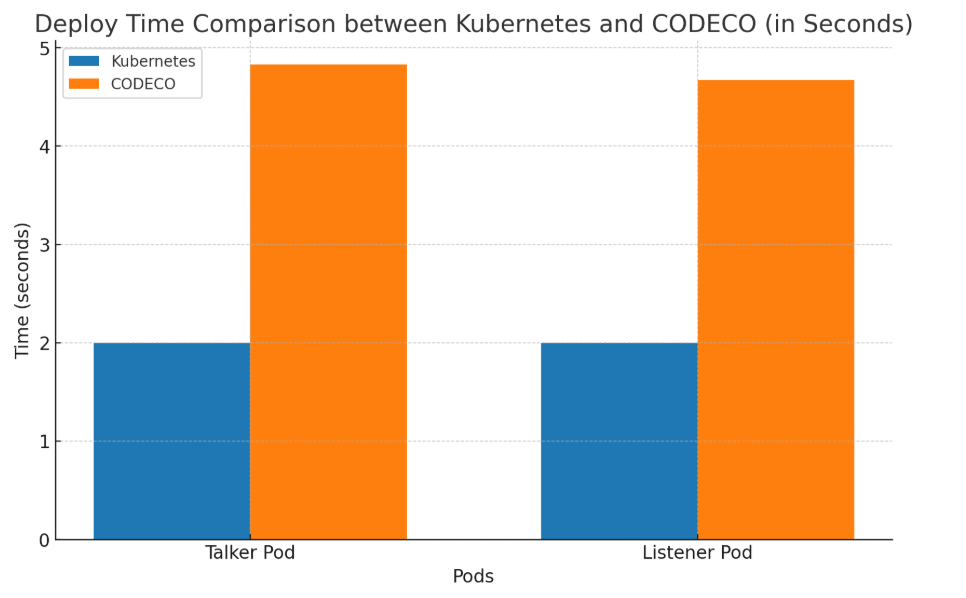}
    \caption{Pod deployment time under Kubernetes and Kubernetes+CODECO.}
    \label{fig:deploy_time}
\end{figure}

\begin{figure}[ht]
    \centering
    \includegraphics[width=0.7\columnwidth]{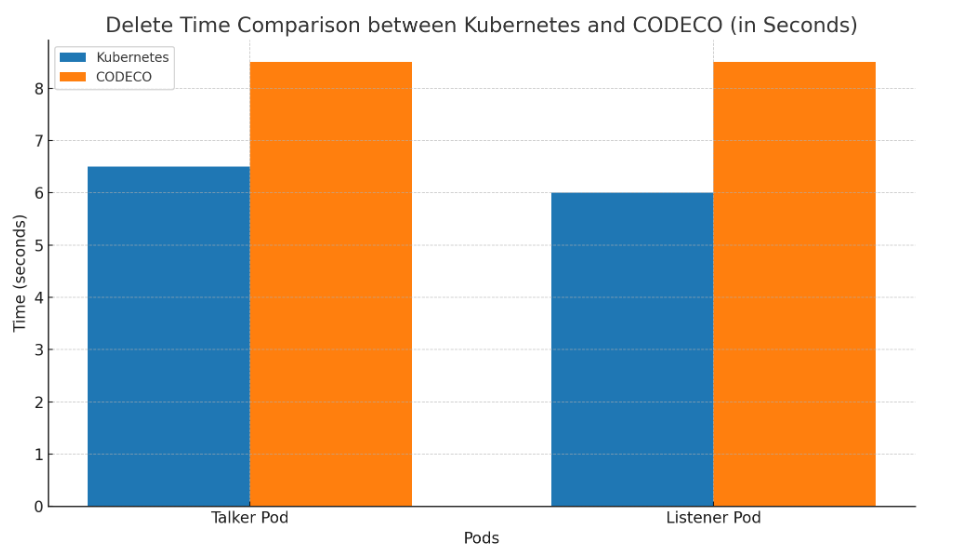}
    \caption{Pod deletion time under Kubernetes and Kubernetes+CODECO.}
    \label{fig:delete_time}
\end{figure}

As shown in Figures~\ref{fig:deploy_time} and~\ref{fig:delete_time}, Kubernetes performs both the deployment and deletion operations faster (approx.\ 2\,s) than CODECO (approx.\ 5\,s). The additional time stems from the initialization and teardown of the VXLAN-based secure overlay (L2S-M\footnote{https://gitlab.eclipse.org/eclipse-research-labs/codeco-project/network-management-and-adaptation-netma/secure-connectivity}) used by CODECO to create a secure overlay for pod-to-pod communication.

While the difference is measurable, both delays are significantly small, given that Kubernetes  lifecycles are in the range of minutes. For latency-critical applications, further optimization may be desirable, and planned real-device tests will explore this impact.

\subsection{Summary of Findings}

The evaluation highlights distinct trade-offs:

\begin{itemize}
\item \textbf{CPU efficiency:} CODECO reduces CPU load vs.\ vanilla Kubernetes.
\item \textbf{Memory:} CODECO introduces a small memory overhead (\textasciitilde10--15\%).
\item \textbf{Networking:} CODECO yields smoother, controlled communication rates.
\item \textbf{Lifecycle latency:} Kubernetes starts/stops pods faster (\textasciitilde2\,s advantage).
\end{itemize}

Overall, Kubernetes provides faster lifecycle operations, while CODECO enhances resource awareness and communication stability. These features are relevant to mobile, resource-constrained AMR environments. These results represent an initial validation which is important in terms of demonstration of capability, and in terms of understanding if CODECO components need to be fine-tuned for AMR environments (as is the case of the networking component).

\section{Related Work}
Containerization and orchestration for robotics have gained significant attention in recent years. Lumpp et~al.\ propose a graph-partitioning algorithm that groups ROS~2 nodes into containers to minimize inter-container communication and reduce memory usage \cite{lumpp2021container}. While effective for static deployment optimization, this approach does not address runtime network variability, robot mobility, or dynamic workload adaptation. In contrast, CODECO supports QoS-aware orchestration, stateful migration, and overlay networking (L2S-M), enabling continuous operation in distributed and mobile robotic environments.

In subsequent work, Lumpp et~al.\ introduce \emph{RT-K3S}, a real-time extension for K3s that incorporates deadline-, period-, and criticality-aware scheduling policies \cite{lumpp2022containerization}. RT-K3S significantly reduces deadline misses compared to native K3s, though it also highlights container-induced overhead in real-time workloads. RT-K3S focuses on single-cluster real-time scheduling, whereas CODECO addresses broader dynamic orchestration challenges across heterogeneous compute nodes, including Edge and AMRs.

RT-Kube \cite{10323202} extends Kubernetes with mixed-criticality scheduling and runtime monitoring, enabling real-time migration of non-critical workloads to preserve resources for critical tasks. While highly effective within a single cluster, RT-Kube does not address distributed mobility or wireless constraints. CODECO complements mixed-criticality scheduling with fleet-aware coordination, stateful micro-service migration, and network-aware decision-making across edge-robot fabrics.

De~Marchi et~al.\ propose ROS4K, which automatically selects communication mechanisms (shared memory, zero-copy, standard DDS) based on system topology \cite{de2024orchestration}. Their work improves transport performance in static deployments. CODECO is complementary: beyond communication optimization, it orchestrates where computation executes under changing energy, mobility, and network conditions, while preserving application state.

Lampe et~al.\ present RobotKube, a Kubernetes-based DevOps framework for event-driven management of robotic micro-services \cite{lampe2023robotkube}. RobotKube supports reactive adaptation within single-cluster deployments. CODECO extends this paradigm with AI-assisted placement, QoS-driven scheduling, multi-cluster Edge–AMR coordination, and stateful migration, enabling robust and resilient operation in dynamic indoor industrial environments.

\section{Conclusions and Future Work}
\label{conclusions}
The case study and respective experimental evaluation demonstrates clear trade-offs between baseline Kubernetes and the CODECO-enhanced orchestration approach for AMR workloads. Baseline Kubernetes achieves slightly faster pod deployment and deletion times and marginally lower memory use, answering RQ2 regarding operational costs. These differences are small in absolute terms and primarily reflect Kubernetes' simplified design, which focus on the infrastructure, and only on application auto-scaling needs.

Conversely, CODECO shows benefits aligned with the requirements of AMR systems. It provides lower CPU utilization, lower data rates and improved network predictability via QoS-aware communication control, and resilience under resource variability, addressing RQ1 and RQ3. Such stability is particularly relevant for real-time AMR workloads where predictable compute and network performance is critical. These findings indicate that the trade-offs favor CODECO in dynamic Edge AMR deployments, especially those requiring communication guarantees, energy-awareness, and continuous operation under wireless or mobility constraints.

The modest memory overhead and additional pod-lifecycle latency observed when applying CODECO primarily arises from its secure VXLAN-based L2S-M overlay and enriched orchestration logic. These costs quantify RQ2 and remain acceptable for most AMR workflows, though optimization may be required for latency-critical AMR swarm operations. 

Overall, results validate that CODECO enhances stateful workload orchestration, provides communication stability, and improves resource efficiency under realistic robotic conditions, complementing Kubernetes and supporting mission-critical AMR and IIoT deployments. 
\medskip
\noindent\textbf{Current work.} Ongoing experiments are evaluating CODECO on a pyhsical testbed relying on embedded AMRs, validating that performance gains and overheads observed in emulation translate to real-world operation. These tests specifically target RQ1 (stateful migration continuity) and RQ3 (communication stability under wireless variability and battery constraints).

\medskip
\noindent\textbf{Future work} will extend this study to more comprehensively answer RQ1-RQ3:
\begin{itemize}
    \item \textbf{Scalability:} increasing the number of AMRs and the complexity of the AMR application, to assess migration efficiency and resource usage at scale.
    \item \textbf{Resilience and robustness:} analyzing system behavior under node mobility, and network degradation, including autonomous stateful migration and recovery dynamics.
    \item \textbf{Energy-awareness:} quantifying energy savings and scheduling efficiency under CODECO’s resource-aware scheduling policies compared to Kubernetes.
    \item \textbf{Federation and multi-tenancy:} evaluating CODECO across federated AMR clusters and shared Edge infrastructure, to assess when orchestration trade-offs favor CODECO in large-scale deployments.
\end{itemize}

\section{ACKNOWLEDGMENTS}
This  work  has  received  support  from  the following projects: European Commission Horizon Europe programme, project CODECO, grant number 101092696; SemComIIoT, grant agreement M-0626 , Sino-German Center (SGC), supported by the Deutsche Forschungsgemeinschaft (DFG).

\def\refname{REFERENCES}

  \bibliographystyle{ieeetr}
  \bibliography{references}

\end{document}